\documentclass[letterpaper, 10 pt, conference]{ieeeconf}  

\IEEEoverridecommandlockouts                              
\makeatletter
\let\NAT@parse\undefined
\makeatother

\usepackage[dvipsnames]{xcolor}

\newcommand*\linkcolours{ForestGreen}

\usepackage{times}
\usepackage{graphicx}
\usepackage{amssymb}
\usepackage{gensymb}
\usepackage{amsmath}
\usepackage{breakurl}

\usepackage{url,hyperref}
\hypersetup{
colorlinks,
linkcolor=\linkcolours,
citecolor=\linkcolours,
filecolor=\linkcolours,
urlcolor=\linkcolours}

\usepackage{algorithm}
\usepackage{algorithmic}

\usepackage[labelfont={bf},font=small]{caption}
\usepackage[none]{hyphenat}

\usepackage{mathtools, cuted}

\usepackage[noadjust, nobreak]{cite}

\usepackage{tabularx}
\usepackage{amsmath}

\usepackage{float}

\usepackage{pifont}

\newcolumntype{Y}{>{\centering\arraybackslash}X}

\usepackage[]{placeins}

\usepackage{placeins}

\usepackage{tikz}

\usepackage[framemethod=tikz]{mdframed}

\usepackage{afterpage}

\usepackage{stfloats}

\usepackage{atbegshi}
\newcommand{\handlethispage}{}
\newcommand{\discardpagesfromhere}{\let\handlethispage\AtBeginShipoutDiscard}
\newcommand{\keeppagesfromhere}{\let\handlethispage\relax}
\AtBeginShipout{\handlethispage}

\usepackage{comment}

\title{\LARGE \bf
A Method for Estimating Reflectance map and Material using Deep Learning with Synthetic Dataset
}

\author{Mingi Lim$^{1}$ and Sung-eui Yoon$^{2}$
\thanks{$^{1}$Mingi Lim, Department of Computer Science, 
Korea Advanced Institute of Science and Technology
Email: mingi@kaist.ac.kr
}%
\thanks{$^{2}$Sung-eui Yoon, Department of Computer Science, 
Korea Advanced Institute of Science and Technology
}%
}

\begin{document}

\maketitle
\thispagestyle{empty}
\pagestyle{empty}

\begin{abstract}

The process of decomposing target images into their internal properties is a difficult task due to the inherent ill-posed nature of the problem.
The lack of data required to train a network is a one of the reasons why the decomposing appearance task is difficult. In this paper, we propose a deep learning-based reflectance map prediction system for material estimation of target objects in the image, so as to alleviate the ill-posed problem that occurs in this image decomposition operation. We also propose a network architecture for Bidirectional Reflectance Distribution Function (BRDF) parameter estimation, environment map estimation. We also use synthetic data to solve the lack of data problems. We get out of the previously proposed Deep Learning-based network architecture for reflectance map, and we newly propose to use conditional Generative Adversarial Network (cGAN) structures for estimating the reflectance map, which enables better results in many applications. To improve the efficiency of learning in this structure, we newly utilized the loss function using the normal map of the target object.

\end{abstract}

\section{INTRODUCTION}

In a traditional computer vision tasks, there are several works to do decomposition of a target image into internal properties. These kinds of works are usually called as inverse rendering.\\
The way people can see things in real life or how cameras can take pictures of their surroundings is widely understood by basic science such as physics. In short, when light hits an object in a particular direction, the material properties of the object determine the direction and proportion of the light's reflection, so that it can be recognized or photographed by the contact of such a reflected light with the eye or camera. However, in the opposite case, such as finding the properties of the real object through the result image is a very difficult problem because the same appearance results can be made by a combination of multiple internal properties.\\
To estimate those properties, it is a common practice to assume that one or more properties are generally known or simplified to estimate other properties. For example, traditional approaches to estimate internal properties assume lambertian materials \cite{LDR1}, or point lights \cite{TSB1}.In this way, in order to simplify the problem, shapes are often provided as 3D models that is used for synthesize dataset or restrict the class of shapes.\\
In this work, we extract high quality reflectance maps from images of a target object with complex shapes and specular materials under complex natural illumination. Futhermore, we use this network for estimating BRDF parameters, environment map, and classifying materials.\\
Based on a previous work, we change several things to make improvements. At first, we introduce cGAN for estimating better reflectance maps. Next, we use our novel loss function using object normal for estimating more exact reflectance map. Finally, we do several applications using estimated reflectance maps such as estimating BRDF parameters, environment maps and classifying materials.\\
A reflectance map maintains the orientation-dependent appearance of a fixed material under a certain surrounding illumination. In other words, a reflectance map is an image of an object's material itself, mixed with the surrounding environment map. By separating this reflectance map, we can obtain an information that we want, material and illumination. Under assumptions such as single material, shadow, distant light source and distant viewer, the relationship between surface orientation and appearance is fully explained by the reflectance map. It can represent all surrounding illuminations and material properties. In particular, it can represent specular materials under high-frequency natural illumination precisely. Therefore, without a better understanding and an analysis of a 2D image system, the ability to estimate reflectance maps comes to a wide spectrum of applications, including estimating BRDF parameters of input objects, environment maps, and classifying materials with additionally synthesized images. The input of our reflectance map estimator is a segmented 2D image where an object from an input image with various shape and the output is a reflectance map.\\
In conclusion, at first, we propose an improved approach that estimates a reflectance map from the input image with deep-learning based cGAN with novel loss using an observable point map. Next, we suggest a model structure that use estimated reflectance map to estimate Cook-Torrance BRDF\cite{CT1} parameters, that is used to render materials of our synthetic dataset and also it is used to estimate environment maps that represent the surrounding illuminations. We make following key contributions:
\begin{itemize}
  \item We propose a deep learning-based method to estimate reflectance maps from 2D images with a cGAN structure and decompose the estimated reflectance map into the material parameters and the environment map.
  \item We suggest a new loss for extracting reflectance maps from input images, using object normal maps.
  \item We introduce a new dataset that includes synthetic images to encourage our training process of our deep-learning sub-models that estimate reflectance maps, environment maps and BRDF parameters.
\end{itemize}

More commonly, destabilizing perturbations are reduced by selecting a low order loss function and stable learning rate. Low order loss functions; such as absolute and squared distances, are effective because they are less prone to destabilizingly high errors than higher-order loss functions. Indeed, loss function modifications used to stabilize learning often lower loss function order. For instance, Huberization \cite{huber1964robust} reduces perturbations by losses, $L$, larger than $h$ by applying the mapping $L \rightarrow \min(L,(hL)^{1/2})$.

\section{Related Works}
The appearance of 3D object relates to its surface geometry, material, and illumination. Estimating these factors is a fundamental problem in a computer vision. After a release of depth sensors, geometry reconstruction have made rapid progress recently. However, the estimation of material and illumination remains relatively more challenging these days. Previous approaches for material and illumination estimation need strong assumptions. For such assumptions, for example, necessity of a depth sensor, multiple images of the same object under varying illumination, limited lighting conditions, a fixed rotation under static illumination, an object of a given class, or requiring user's intervention.\\

\textbf{Reflectance Maps} Reflectance maps assign a surface appearance to its orientation, therefore it is a combination of material reflectance and surrounding illumination that can be used in many important applications. For example, if we have 3D models, we can render virtual 3D objects with the same material and the illumination of a real existing object or change the appearance of two objects\cite{RSD1}. In computer graphics, there are many ways to represent object materials, but a reflectance map is a one of the popular way to represent targets' material with their surrounding environments easily. In this work, we want to solve challenges of the related previous work, Georgoulis et al. \cite{RSD1} with estimating reflectance map more exactly using conditional GAN \cite{GA1}. These estimated reflectance maps can be used in many applications and make the results better, such as BRDF parameter estimation, environment map estimation, material classification.\\

\textbf{Material Estimation} Depending on the assumption of a geometry, we can divide the material estimation area into two main categories. At first, there are approaches that assume geometry are known. Methods that require the surface geometry of objects to be known can work on any type of surface geometry when object geometry is provided. Dong et al. \cite{AFM1} estimate spatially-varying reflectance from the video of a rotating object of known geometry. Wu and Zhou \cite{AF1} and Knecht et al. \cite{IBE1} propose a method for appearence estimation at on-the-fly frame rates using a Kinect sensor. Li et al. \cite{MSA1} targets for estimating planer object's surface appearance from single images using self-augmented CNNs. And next, there are approaches for specific object classes of unknown geometry. There are some recent works, Rematas et al. \cite{DRM1}, Georgoulis et al. \cite{RSD1} and Liu et al. \cite{ME1} 
Unlike works described earlier, these three works can estimate materials without the need for additional sensors to obtain geometry. However, for these works, the actual available target's shape is limited to the shape class of the object used for training. Therefore, their work only works on certain types of objects, such as cars and chairs. By contrast to these methods, our approach uses general surface geometry so do not restricted to specific object classes.

\section{Backgrounds}

Before going into explanation of our methods, we introduce some basic definitions that can be used in our paper. We can represent reflectance map $L \left( \omega \right) \in S^{+} \rightarrow  R^{3}$ which is a map from orientations $\omega$ in the positive half-sphere $S^{+}$ to the RGB radiance value L leaving that surface to a distant viewer. It combines the effect of reflectance and illumination.\\ 
Horn and Sjoberg \cite{CR1} suggest to use positional gradients to parametrize orientation $\omega$. Instead, Georgoulis et al. \cite{RSD1} parametrize the orientation simply by s and t those are normalized surface normal’s x and y components. We follow their approach to parameterize a reflectance map. They also drop the z coordinate which is equivalent to drawing a sphere under orthographic projection and it can be represented as a single image. Note that this only covers the upper half-sphere $S^+$, so they need to parametrize a half-sphere.\\ 
We explain the rendering equation first to approach the concept of the reflectance map. Let's start with the rendering equation below.\cite{RE1}.
\begin{multline}
    L_{o}(x, \omega_{o}) = L_{e}(x, \omega_{o}) + \\ \int_{\omega^{+}} f_{r}(x, \omega_{i}, \omega_{o})L_{i}(x, \omega_{i}) \left< \omega _{i}, n(x) \right>^{+} d\omega _{i}
\end{multline}
where $L_o$ is the outgoing radiance, $L_e$ the emitted radiance, $L_i$ the incoming radiance, $f_r$ the BRDF, and $n(x)$ the surface orientation. Radiances can be represented as functions of position x and direction $\omega$. The reflection part is the integral over the upper hemisphere $S^{+}$ of the product of incoming light $L_{i}$, BRDF $f_r$, and the dot product of surface normal n(x) and integration direction $\omega_{i}$. 
Refer to a recent book on the rendering~\cite{YOON18} for more detailed information.\\
In this work, there are several assumptions. We only consider 1) general materials without emissive, translucent materials and 2) a single material which means one surface reflectance model. We also consider that 3) the object is seen under orthographic projection from an infinitely far-away observer and 4) illuminated under image-based lighting technique with no shadows. These assumptions can make the rendering equation the following function:
\begin{multline}
    L_{o}(\omega_{o}) = \int_{\Omega^{+}} f_{r}(\omega_{i}, \omega_{o})L_{i}(\omega_{i}) \left< \omega _{i}, n \right>^{+} d\omega _{i}
\end{multline}
where $L_{o}$ is the reflectance map, $L_{i}$ is the illumination, and $f_{r}$ is the surface reflectance. In addition, we refer to the surface reflectance $f_{r}$ as the material for simplification. 
We suggest to use Cook-Torrance Model for rendering realistic, physical based material used widely. \cite{CT1}: 
\begin{align} 
    f_{r}=k_{d}f_{lambert}+k_{s}f_{cook-torrance}, \\ 
    f_{lambert}=\pi / c,
\end{align}
where $k_{d}$ is called the diffuse color, $k_{s}$ the specular color. We use the lambert reflection model to model a diffuse reflection.
\begin{multline}
    f_{cook-torrance}\left( \omega_i, \omega_o \right) = \frac{ DFG }{ \left(n \cdot \omega_{i} \right) \left( n \cdot \omega_{o} \right)}.
\end{multline}
Cook-Torrance specular model consists of 3 different functions that models the laws of physics in real world, D stands for normal distribution function, G stands for geometric shadowing function, and F stands for Fresnel function. The formula for each parameters is as follows:
\begin{align}
    D_{GGX} \left( m \right) = \frac{ \alpha ^{2} }{ \pi \left( \left( n \cdot m \right)^{2} \left(\alpha^{2}-1 \right) +1 \right) ^{2} } \\
    G_{GGX}  \left( v \right) = \frac{ \alpha ^{2} }{ \left( n \cdot v \right)   + \sqrt{ \alpha^{2} +  \left( 1 - \alpha^{2} \right) \left( n \cdot v \right)^{2} } }
\end{align}
\begin{align}
    F = F_{0} + \left( 1 - F_{0} \right) \left( 1 - cos\left( \theta \right) \right)^{5} \\
    F_{0} = \left( \frac{ n_{1} - n_{2} }{ n_{1} + n_{2} } \right).
\end{align}
As both the illumination $L_{i}$ and the reflectance map $L_{o}$ are two-dimensional functions of direction $\omega$, we represent them as images using s, t that we explained above.

\begin{figure*}
  \includegraphics[width=\linewidth]{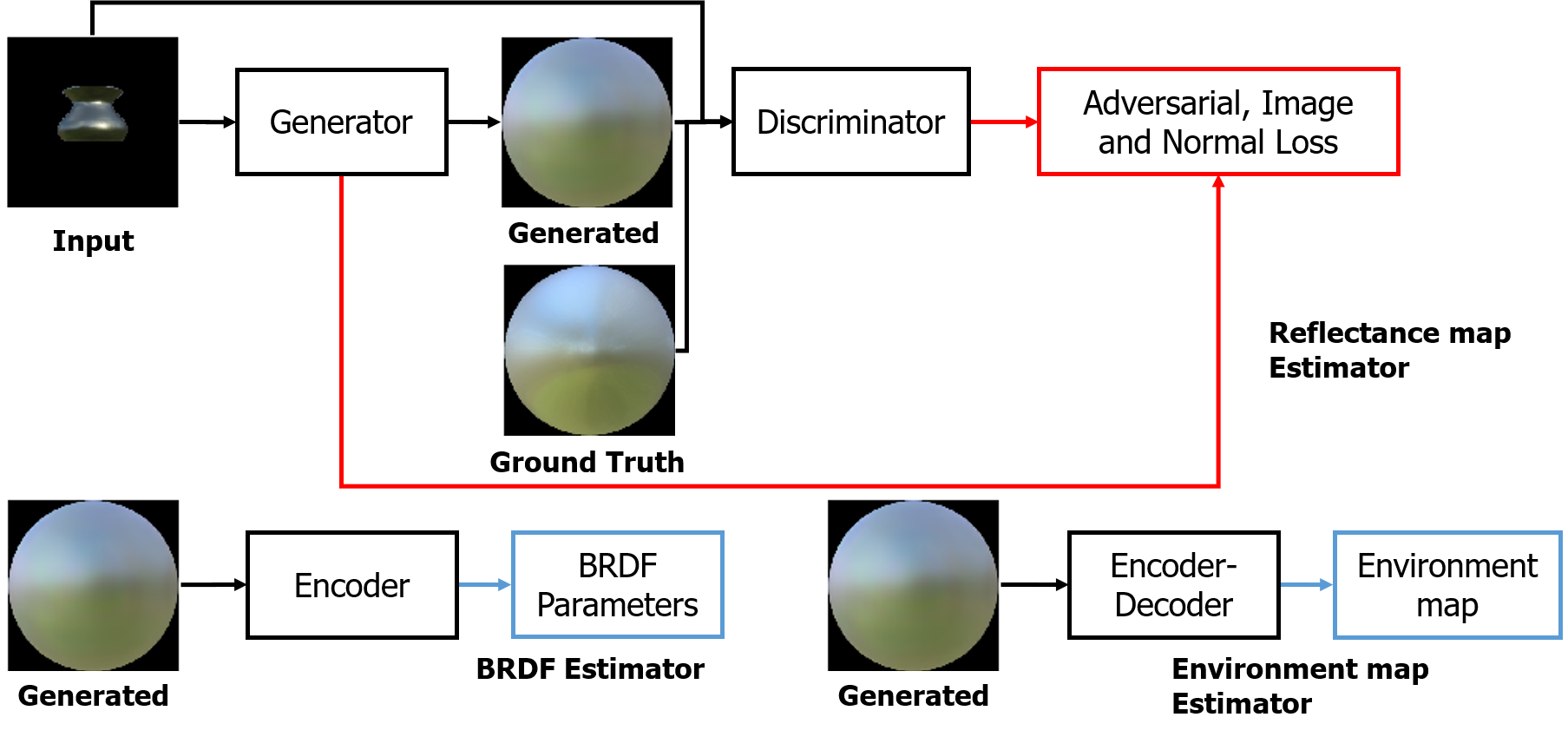}
  \captionof{figure}{Overview of Network Architectures.}
  \label{fig:all_networks}
\end{figure*}

\section{Network Architectures}
We devise network architectures to accomplish our goals as shown in Fig. \ref{fig:all_networks}. There are 4 separate networks. First, there is a network that estimates the reflectance map from an input image. We opt for cGAN~\cite{GA1} to estimate the reflectance map more precisely and design a new loss to help the network train well.

\subsection{Network for Reflectance Map}

\begin{figure}[tbp]
  \includegraphics[width=\linewidth]{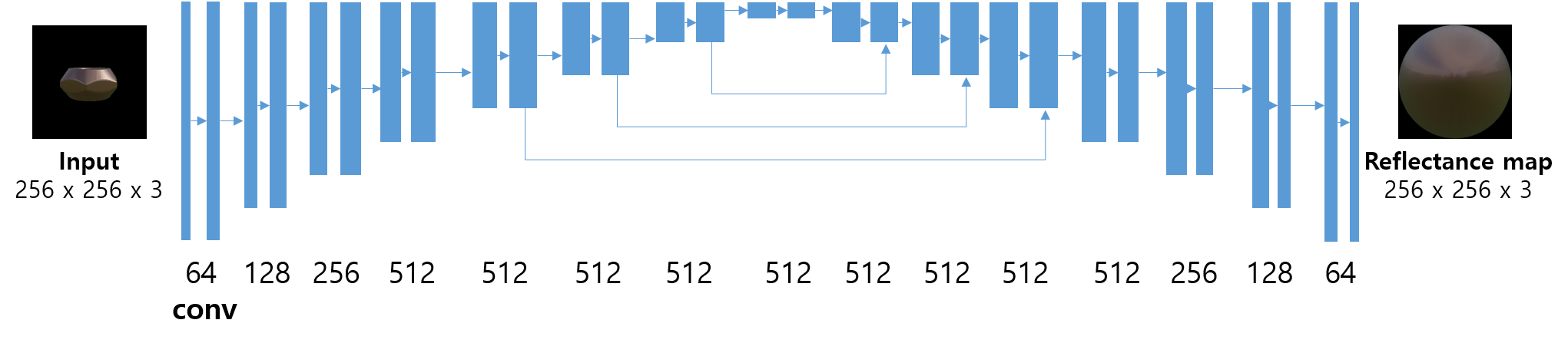}
  \captionof{figure}{A generator model architecture for estimating a reflectance map. The bottom numbers represent the number of channels of the above convolutional layers.}
  \label{fig:all_networks}
\end{figure}

\begin{figure*}
  \includegraphics[width=\linewidth]{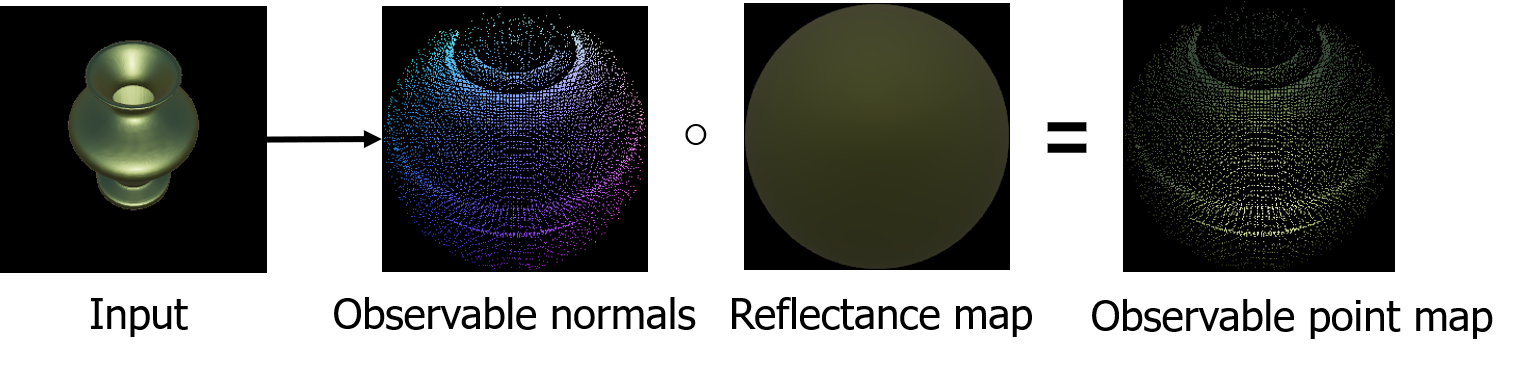}
  \captionof{figure}{Overview of calculating normal loss for training reflectance estimator.}
  \label{fig:normal_loss}
\end{figure*}

We want to estimate the reflectance map of a target object in an input image. A target object is consist of a single material and an input image is 2D RGB image. This is equivalent to estimating an appearance of sphere with the same material as the target object would look like from the same camera position and under the same illumination. Previous work uses an encoder-decoder network with L2 loss~\cite{RSD1}. We want to improve the quality of the reflectance map itself with expecting performance improvement of other applications. Therefore we used cGAN with a loss using object normal for estimating reflectance map. 
There are several works using cGAN for transforming images.  We first deploy to extract reflectance map from input images and then we applied some changes appropriate for estimating reflectance map. First, we do not use patches for estimating a reflectance map (known as PatchGAN~\cite{ITC1}). Because there are no patch-to-patch relations between input and output images, we adopt our novel loss using objects' normal for estimating a more realistic and exact reflectance map. \\
Also, our deep-learning networks are trained and evaluated using our new synthesized dataset. In our approach, our network for estimating reflectance map learn a mapping between an input image and its reflectance map, following a cGAN architecture.\\
Fig. \ref{fig:all_networks} shows the proposed architecture. We introduce half-connected U-Net architecture. In the beginning, there are several convolutional layers with batch normalization, ReLU, and pooling layers, the size of the input feature maps is reduced to 1 x 1. Next on, there are several up-convolutional layers until the output size becomes 256 x 256 pixels. In 3 low-level layers, each incomming input features is concatenated with outcomming CNN features. Every convolutional layer has a one stride and zero paddings where the output can be the same size as the input. L1 loss between the ground-truth reflectance map and prediction is used in the final layer. Furthermore, this generator is trained with adversarial loss calculated using real/fake classifier and loss with observable point map.\\
In short, the intrinsic goal of this network is find a mapping function from the input image to a reflectance map. This model is going to learn not only how to map the image pixels to locations on the reflectance map, but also to interpolate appearance from observed normals. Our goal is a challenging problem as a kind of domain transfer problem that encodes image to a directional domain.
In the previous approach\cite{RSD1}, they used encoder-decoder structure, which actively utilizes CNN structure, is used for training. Furthermore, they use a sparse reflectance map as one of inputs, an image that records material appearances for observable normal orientation, and a general object image is used as an another input. As such, they try several ways, but do not show much difference in results between different two inputs. Inspired by this, we want to use cGAN to improve the quality of the result image more.\\
And then, we will explain our loss for the training network for reflectance estimation. At first, the objective of conditional GAN \cite{GA1} can be expressed as:
\begin{multline}
    {L_{cGAN}}(G,D) ={E_{x,y}}[logD(x, y)]+ \\ {E_{x,z}}[log(1 - D(x,G(x, z))],
\end{multline}
where G tries to minimize this objective against an adversarial D that tries to maximize it. 
Various previous studies have often shown good results using a combination of a GAN loss function and a traditional loss function. The role of discriminator is decision which the target is real or fake, and the generator should not be only trained in the direction of deceiving the discriminator, but also in a way that is substantially less different with pixel values. In this case, the L1 loss function results in a clearer image than the L2 loss function according to the previous work.\cite{ITC1}
In this work, based on good results from previous works, bringing better results by using an additional loss function.

\begin{equation}
    L_{L1}(G) = E_{x,y,z}[||y - G(x, z)||_{1}].
\end{equation}
We then use the ideal object normal to define an observable normal map of an input object. The estimated reflectance map is derived from observable surfaces of the target object therefore unobservable areas of the reflectance map are just estimated from those observable surfaces. Therefore, we propose a new loss to weight on the observable area of the target object as below:
\begin{equation}
    L_{N}(G) = E_{x,y,z}[||(y - G(x, z)) \circ N ||_{1}].
    \label{eq:normal_loss}
\end{equation}
where $N$ is an observable point map of the input object, and $\circ$ means a pixel-wise product between images. You can see details in Fig. \ref{fig:normal_loss}. 
We call this loss as normal loss because it uses observable normal of input object for calculating loss. An observable normal map is a sphere map of observable object normal orientation. It is a mapped image from an object normal to an orientation similar to the mapping method of a reflectance map. An observable point map is a sphere map of observable points' pixel value with their orientation. we obtain an observable point map by a pixel-wise product with binarized observable normal map and the estimated reflectance map. The observable point map is a map that has pixel values on each orientations for observable points on the reflectance map and an expectation for the observable point map is used to train.

Therefore, our final objective is:
\begin{multline} \label{eqn_fin}
    G* = arg min_{G} max_{D}L_{cGAN}(G,D) + \\ \lambda _{1}L_{L1}(G) + \lambda_{2}L_{N}(G).
\end{multline} 

\subsection{Network for Environment Map and BRDF Parameters}

\begin{figure}[tbp]
  \includegraphics[width=\linewidth]{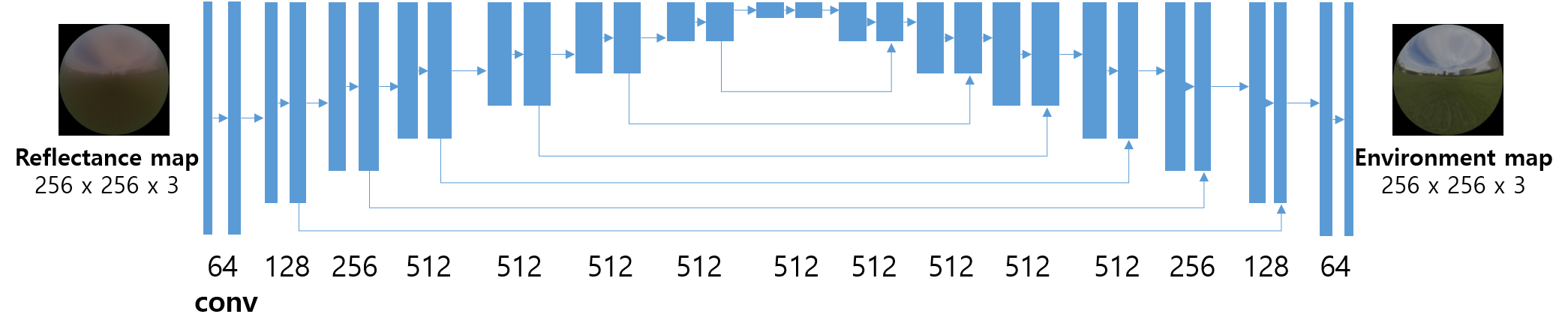}
  \captionof{figure}{Architecture of the model for an environment map estimation. 
  Bottom numbers represent sizes of the feature channels of the corresponding convolutional and deconvolutional layers.}
  \label{fig:envmap_estimator}
\end{figure}

\begin{figure}[tbp]
  \includegraphics[width=\linewidth]{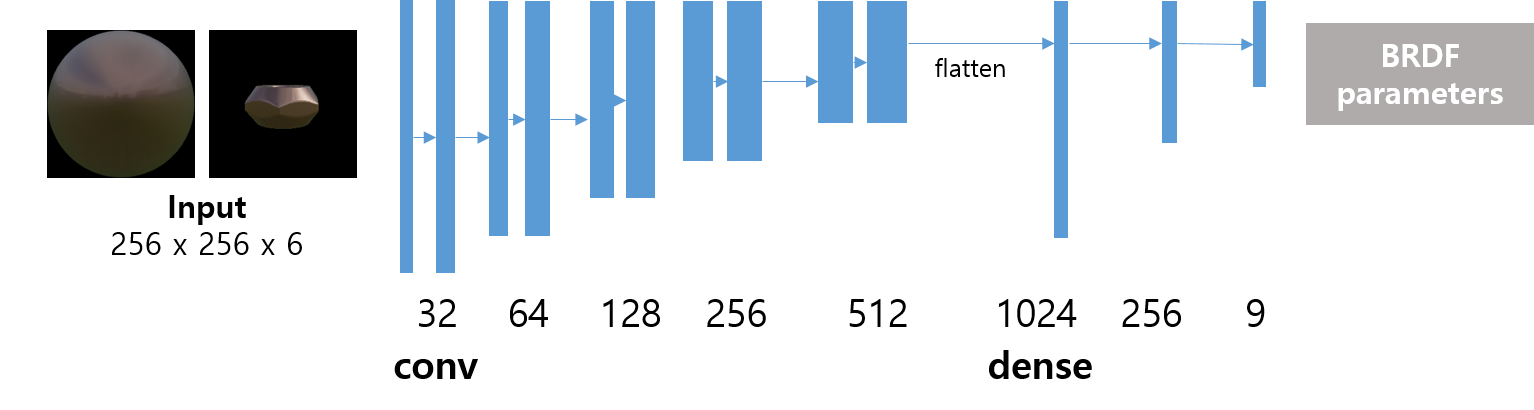}
  \captionof{figure}{Architecture of the model for a BRDF estimation. Bottom numbers represent sizes of the feature channels of the corresponding convolution and dense layer.}
  \label{fig:brdf_estimator}
\end{figure}

In the previous section, we showed our model and training method for estimating a reflectance map of a target object from a RGB input image. In this section, we are going to explain how to decompose the estimated reflectance map into its internal properties: materials and illumination.\\
The input of our material and illumination decomposition is a reflectance map estimated from the reflectance map estimator. Commonly, a reflectance map can be obtained in several ways. For example, when a spherical sample of the real material is available, it can be obtainable directly with surrounding illumination. In addition, in the case of different shape, if the shape is known, more specifically, its normals are known, its reflectance map can be retrieved, at least for all observed surface orientations. By the way, If the shape is unknown, there are several alternatives that have been tested to acquire it such as 3D scanning, structure-from-motion, depth sensors, CNN-based depth extraction \cite{EIGEN14} \cite{LI15} \cite{LIU15} or directly estimating the normals using deep learning\cite{EIGEN15}. In this paper we going to use estimated reflectance map, that is given from our first network, reflectance map estimator.
Our outputs explained in this section are Cook-Torrance BRDF parameters, and an environment map. 
The environment map is rendered using HDR spherical image, expressing surrounding illumination. We show details of the networks in Fig. \ref{fig:brdf_estimator} and Fig. \ref{fig:envmap_estimator}. All networks take a reflectance map as input. 
BRDF estimator make a 4-dimensional parameter vector, which are corresponding to each parameter of the Cook-Torrance reflectance model: three color values for the specular, and a roughness value, which defines how rough the material is. The environment map estimator estimates the environment map.\\ For BRDF estimator network, we use Huber loss~\cite{HUBER64} for regression and for the environment map estimator network, we use L2 loss. We also decide the resolution of the output of the environment map estimator as 256 x 256, which is more improved quality comparing to the previous work.
The input type of this network is a 2D image of the reflectance map and the output type is Cook-Torrance parameter vector. The design of the network is shown in Fig. \ref{fig:brdf_estimator}. \\
Overall, the environment map estimator consists of multiple convolutional layers, and deconvolution layers. Those convolutional layers have batch normalization and ReLU. Next, there are several fully-connected layers for parameter regression.
As we already mentioned, an input of these networks is the same reflectance map as in the BRDF estimator. We opt for U-Net architecture to estimate an environment map with preserving what input features have. As other model previously mentioned, these models apply ReLU and batch normalization in each CNN layer units.



\section{Synthetic Dataset}

\begin{figure}[tbp]
  \includegraphics[width=\linewidth]{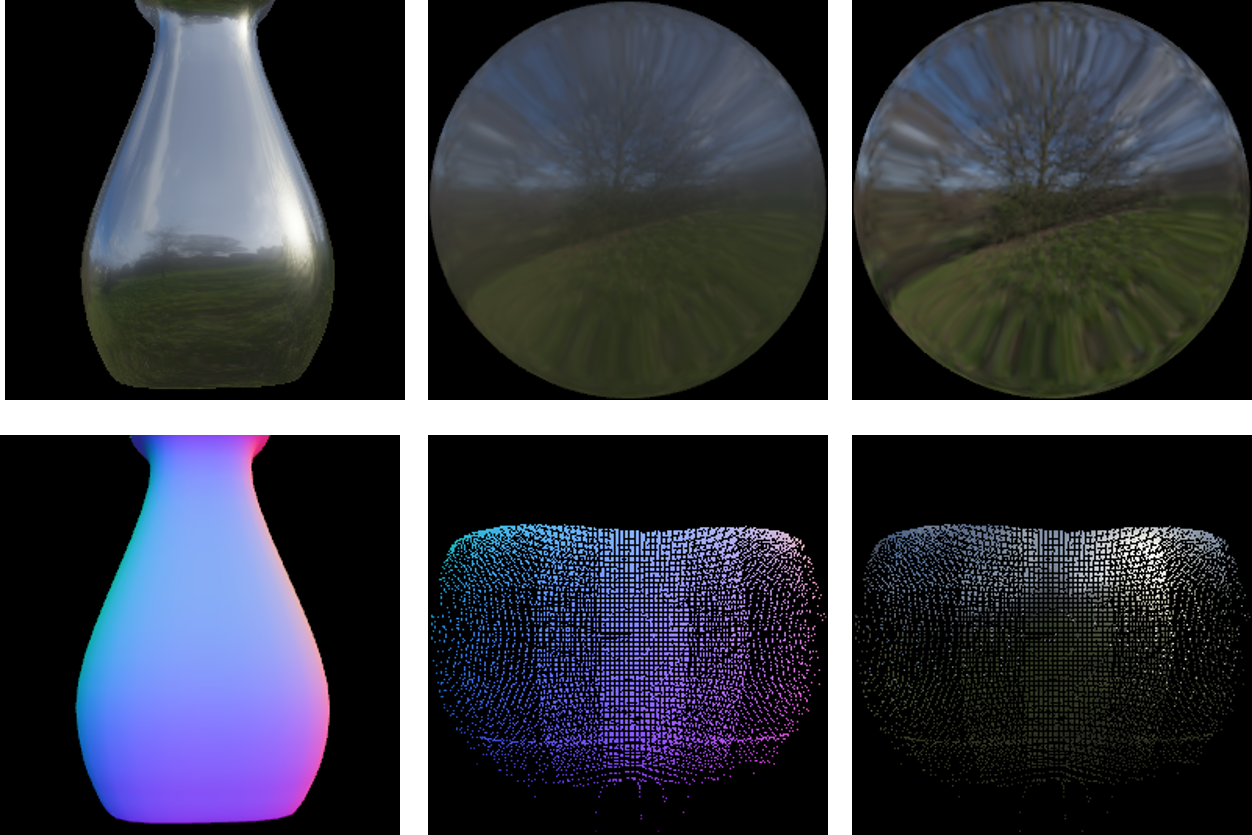}
  \captionof{figure}{Our dataset for estimating the reflectance map, the environment map and BRDF parameters. \\
  From the top-left, an input image, a reflectance map, an environment map, an object normal, an observable normal map, an observable point map of input object.}
  \label{fig:dataset}
\end{figure}

A lot of images are required to train our network. But it is very difficult to find a large dataset. We need a dataset contains many real images with 3D shape, material, and HDR illumination that can be used as ground-truth. Therefore, we consider to synthetize imaginary rendered images for the training process. And also, there is another problem which is a lack of large databases of real material samples and HDR illumination maps. Accordingly, we generated a dataset for training each networks. It includes a set of input images, reflectance maps, environment maps, object normal images, observable normal maps, observable point maps. We also synthesized some results to train different methods for baseline comparison.\\
Synthetic images were produced with random 1) views, 2) 3D shapes, 3) materials and 4) illumination. An example can be seen in Fig. \ref{fig:dataset}. This example was sampled from a random position around the object, looking at the center of the object. We could get 60 3D shapes from the free repository\cite{3DREPO}. For each sample, the object orientation and camera view around the y-axis were selected randomly. Illumination was provided by 61 free HDR illumination maps collected from the internet\cite{ENVREPO}. For materials, we decided to randomize parameters. Overall 107 k sample images were generated. We defined a training-test split where no shape, material or illumination is shared between train and test dataset.

\section{Experimental Setup}
We propose a network to estimate target images from input images with an end-to-end architecture using conditional GAN. Because we want to improve a performance of the previous work \cite{RSD1} and the performance of indirect translation from an input to a reflectance map of the existing work is not significantly improved from the direct method, we suggest the method using the input image directly without estimating normal. This result is validated under our synthesized dataset. 
Let me explain a training process. We train our reflectance map estimator using the synthetic dataset and next, we train the rest of networks using the same dataset. We train BRDF estimator and environment map estimator with results obtained by reflection map estimator as inputs. In the following sub-sections, we share quantitative results obtained using our synthetic dataset, and then figure out how much performance there has been over the traditional method. For rendering our test and train dataset, 3D models, camera pose, rotation of objects, BRDF parameters, and environment map are randomly set and used, as described in the previous section, and there is no sharing of 3D models and environment maps between test dataset and train dataset.

\section{Results}
\begin{table*}
    \begin{center}
        \begin{tabular} {cccccc}
            \hline\hline
            & Methods & DSSIM & MSE & MAE &\\
            \hline
            & \textbf{Our method} & \textbf{0.3248} & \textbf{0.1223} & \textbf{0.2514}&\\
            \hline
            & Ours w/o normal loss & 0.3516 & 0.1455 & 0.2808&\\
            \hline
            & ReflectanceNet\cite{RSD1} & 0.4719 & 0.3377 & 0.4706&\\
            \hline
            & Half-connected U-Net & 0.4087 & 0.2296 & 0.3697&\\
            \hline\hline
        \end{tabular}
    \end{center}
    \captionof{table}{Results of different methods}
    \label{tbl:Result}
\end{table*}

\begin{table*}
    \begin{center}
    
        \begin{tabular} {cccc}
            \hline \hline
            Task & Huber & MSE & MAE\\
            \hline
            Material estimation & 0.0223 & 0.0446 & 0.1650\\
            \hline \hline
        \end{tabular}
        
        \bigskip
        
        \begin{tabular} {cccc}
            \hline \hline
            Task & DSSIM & MSE & MAE\\
            \hline
            Environment map estimation & 0.3092 & 0.0855 & 0.2052\\
            \hline \hline
        \end{tabular}
        
    \end{center}
    \captionof{table}{Results of material estimation, and environment map estimation}
    \label{tbl:Result2}
\end{table*}

\textbf{Quantitative Results}
Our quantitative results are summarized in Table \ref{tbl:Result}. 
We quantitatively analyze our method’s performance to validate our model designs. We compare each networks with our test dataset that contains 15,300 synthetic test images. We compare our approach to three different methods using the same training set. We use three different metric for comparing each different methods such as DSSIM, MSE, and MAE. Structural dissimilarity (DSSIM)\cite{SSB1} is an image distance metric, that corresponds better to the human perception than MAE or RMSE. Mean Squared Error(MSE) measures the average of the squares of the errors that is, the average squared difference between the estimated values and the actual value. Mean Absolute Error (MAE) is the average distance between each pixel point.\\
We also test different methods. 1. Our network as-is, but without the novel normal loss (Eq. \ref{eq:normal_loss}). Exclusion of the normal loss leads to reduced accuracy in the reflectance map estimates. 2. This method uses encoder–decoder structure of Georgoulis et al.\cite{RSD1} using our synthetic training dataset, and estimates a reflectance map of size 32 x 32. This approach performs poorly comparing to other methods. We suspect this might be due to an encoder-decoder model which results in the loss of high-frequency information rather than our cGAN based model with additional losses. 3. We bring the previously mentioned generator structure of reflectance map estimator, half-connected U-Net, which consist of encoder-decoder structure. We tried to construct a training method similar to previous works using L2 loss rather than using our conditional GAN based structure or normal loss suggested in this paper. We train those all networks until convergence. We can find that it records higher loss than our method, that is, we can infer that our methods produce better results.\\
We also report the accuracy of our sub-network for estimating an environment map and BRDF parameters in Table \ref{tbl:Result2}. We estimate BRDF estimator using Huber loss function. Huber loss function \cite{HUBER64} describes the penalty incurred by an estimation procedure. We want to show that this application sub-modules are well trained using an estimated reflectance map. \\

\begin{figure*}
  \includegraphics[width=\linewidth]{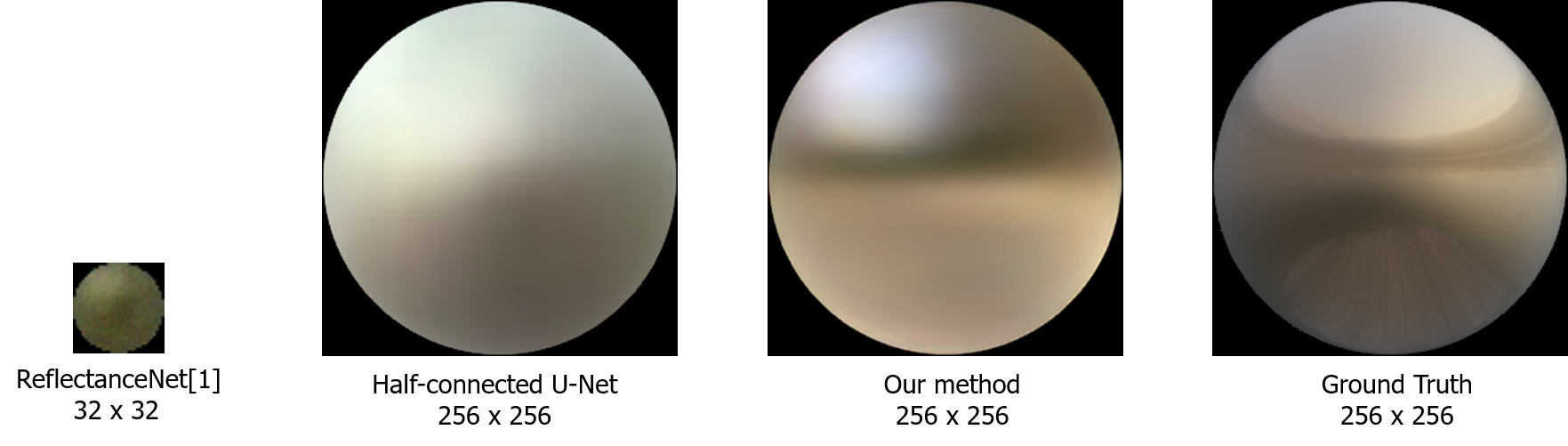}
  \includegraphics[width=\linewidth]{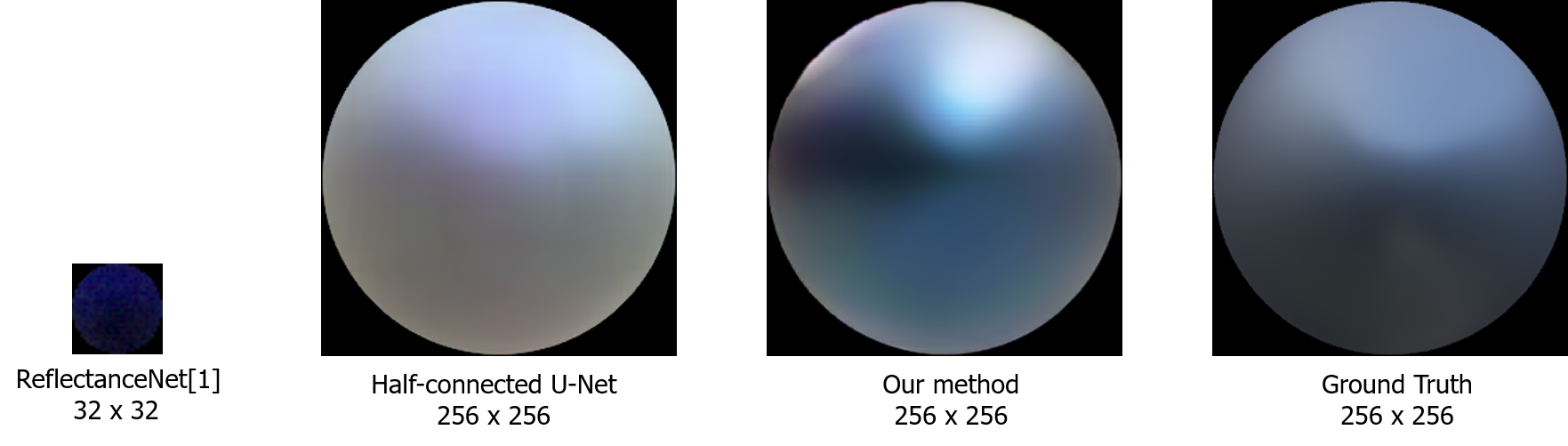}
  \captionof{figure}{Comparison of a reflectance map estimation results based on a single input image. Our approach estimates improved results comparing to other methods}
  \label{fig:examples1}
\end{figure*}

\begin{figure*}
  \includegraphics[width=\linewidth]{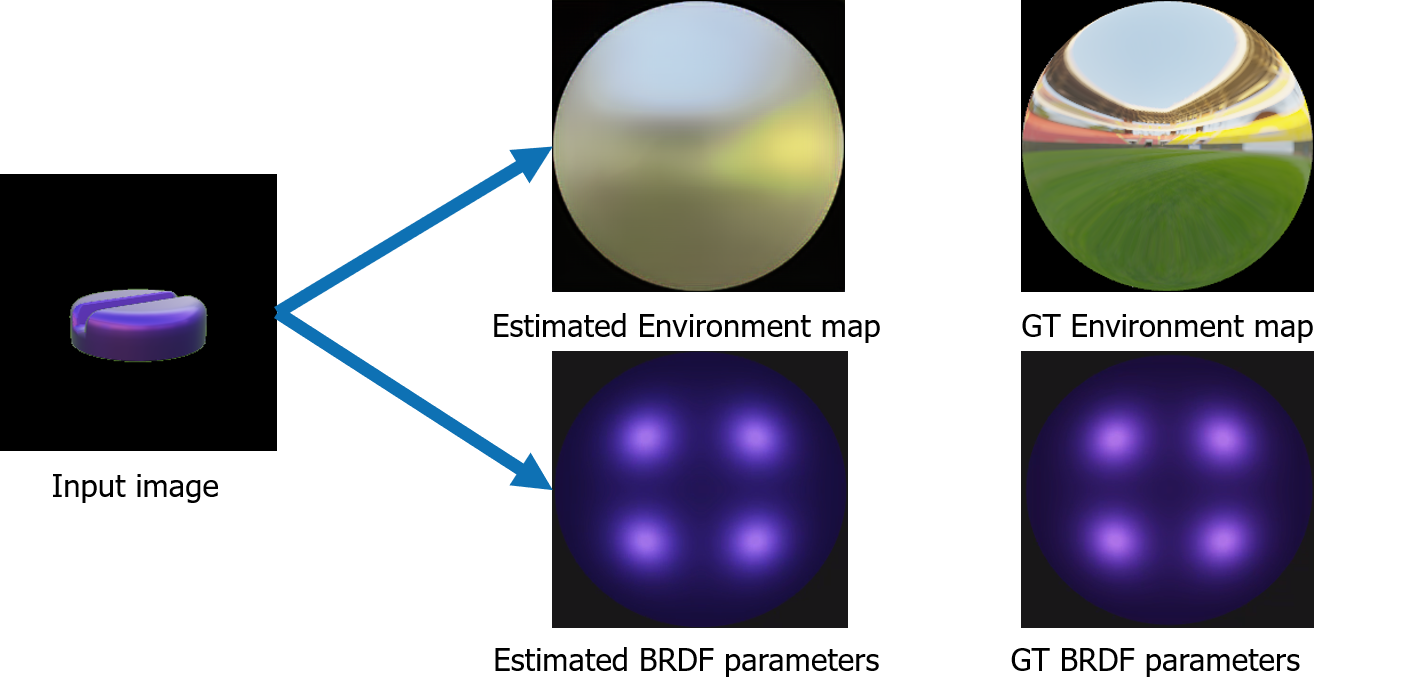}
  \captionof{figure}{Results of environment map estimation and visualization}
  \label{fig:examples3}
\end{figure*}

\textbf{Qualitative Results}
Figure \ref{fig:examples1} shows material estimation results for various methods with same materials and general objects. As you can see, our approach estimates a reflectance map at high quality. 
To take a closer look, ReflectanceNet is a network proposed by the existing work\cite{RSD1}, while the Half-connected U-Net is an encoder-decoder model that is trained by L2 Loss. To train that ReflectanceNet, we resized the dataset synthesized in this work to match the size of the target network, while retaining the structure proposed by the existing work. Comparing results with the ground-truth image, in the case of a ReflectanceNet, the image size is 32x32, and you can see the lack of details. In addition, in the case of Half-connected U-Net, the size of result images is the same as the size of results of our method, you can see that it also lacks details comparing with our method's result.\\
In addition, we are going to show visualization of BRDF parameter estimation results and an environment map in Fig. \ref{fig:examples3}. You can find that there is less difference between visualized estimated BRDF parameters and ground-truth and an example of an environment map is estimated well. Our approach targets on a general shape, and generates high-resolution images. In contrast, previous approaches only work on a limited object class or low-resolution images with blurry results.

\section{Conclusion and Future work}

We have presented our cGAN based material estimation network with estimating a more precise reflectance map, environment map, and classifying materials. We also proposed our new loss for extracting a more exact reflectance map and tackled the lack of data by using a newly rendered synthetic dataset. We have tested our network and method comparing to other networks and verified that it shows some improvement over the prior methods for the estimating reflectance map itself. It can be used for other applications such as BRDF parameter estimation or environment map estimation and improve results by its better quality.\\

However, our work have some limitations. Because our work only utilizes the light reflected from the target object, it targets on a single object and material only. Therefore, it is often unclear to guess surrounding illuminations if the target material is diffuse or if observable normals of the target object are limited. To solve this problem, it would be better to think about how to infer the material of an object by obtaining it in a different way, beyond the estimation of the problem through the objects' appearance only.


\bibliographystyle{ieeetr}


\clearpage

\end{document}